# Reshaping free-text radiology notes into structured reports with generative transformers

Laura Bergomi[a,*] [0009-0006-0359-5128], Tommaso M. Buonocore[a] [0000-0002-2887-088X], Paolo Antonazzo[b], Lorenzo Alberghi[b], Riccardo Bellazzi[a,d] [0000−0002−6974−9808], Lorenzo Preda[b,c] [0000-0002-5479-2766], Chandra Bortolotto[b,c] [0000-0002-9193-9309], Enea Parimbelli[a] [0000−0003−0679−828X]

[a] *Department of Electrical, Computer and Biomedical Engineering, University of Pavia, Pavia, Italy*
[b] *Radiology Unit - Diagnostic Imaging I, Department of Diagnostic Medicine, Fondazione IRCCS Policlinico San Matteo, Pavia, Italy*
[c] *Diagnostic Imaging Unit, Department of Clinical, Surgical, Diagnostic, and Pediatric Sciences, University of Pavia, Pavia, Italy*
[d] *LIM-IA - Laboratory of Medical Informatics and AI, IRCCS Istituti Clinici Scientifici Maugeri, Pavia, Italy*

[*] *Correspondence to: Department of Electrical, Computer and Biomedical Engineering, Via Ferrata 5, 27100, Pavia, Italy. E-mail address: laura.bergomi01@universitadipavia.it (L. Bergomi)*

## Abstract

BACKGROUND: *Radiology reports are typically written in a free-text format, making clinical information difficult to extract and use. Recently the adoption of structured reporting (SR) has been recommended by various medical societies thanks to the advantages it offers, e.g. standardization, completeness and information retrieval. We propose a pipeline to extract information from free-text radiology reports, that fits with the items of the reference SR registry proposed by a national society of interventional and medical radiology, focusing on CT staging of patients with lymphoma.*

METHODS: *Our work aims to leverage the potential of Natural Language Processing (NLP) and Transformer-based models to deal with automatic SR registry filling. With the availability of 174 radiology reports, we investigate a rule-free generative Question Answering approach based on a domain-specific version of T5 (IT5). Two strategies (batch-truncation and ex-post combination) are implemented to comply with the model's context length limitations. Performance is evaluated in terms of strict accuracy, f1, and format accuracy, and compared with the widely used GPT-3.5 Large Language Model. A 5-point Likert scale questionnaire is used to collect human-expert feedback on the similarity between medical annotations and generated answers.*

RESULTS: *The combination of fine-tuning and batch splitting allows IT5 to achieve notable results; it performs on par with GPT-3.5 albeit its size being a thousand times smaller in terms of parameters. Human-based assessment scores show a high correlation (Spearman's correlation coefficients>0.88, p-values<0.001) with AI performance metrics (f1) and confirm the superior ability of LLMs (i.e., GPT-3.5, 175B of parameters) in generating plausible human-like statements.*

CONCLUSIONS: *In our experimental setting, a smaller fine-tuned Transformer-based model with a modest number of parameters (i.e., IT5, 220M) performs well as a clinical information extraction system for automatic SR registry filling task, with superior ability to discern when an N.A. answer is the most correct result to a user query.*

# 1 Introduction

One of the main challenges of Artificial Intelligence (AI) is the automatic processing of large amounts of unstructured textual data. Natural Language Processing (NLP) is a subfield of AI concerned with the development of algorithms capable of processing, analyzing and understanding large amounts of such data (text or speech) in human language. The application of these methods in the biomedical domain is called Clinical NLP (cNLP) and refers to the analysis of clinical narratives and their manipulation and interrogation. English is by far the most resource rich language that has contributed to the development of the cNLP; on the other hand, its use and subsequent performance evaluation are still limited for other languages, e.g. Italian, due to a lack of available data [1]. Annotations on clinical narratives by medical experts are often necessary to train supervised machine learning algorithms. Unfortunately, a medical observation can be affected by the interpretation, experience and abbreviations used by the specific author [2]. All these considerations constitute key challenges for the current SoA of Natural Language Understanding (NLU).

In radiology, a large amount of textual data is generated daily in the form of free-text reports (e.g., transcriptions). Many medical societies (e.g. European Society of Radiology and Radiological Society of North America [3,4]) recognize the increasing need for the adoption of structured reporting (SR) in clinical practice and encourage all institutions to conceive reference registries. These premises motivated the Italian Society of Medical and Interventional Radiology (SIRM) to design structured reports for CT scans of patients with oncological conditions such as Breast, Lung, Colon, Rectum, Lymphoma, Pancreas, and Stomach cancer, as well as Covid19 [5]. However, the fast rate at which unstructured clinical information is being created calls for NLP solutions to transform existing reports into structured representations [2,6].

## 1.1 Background

### 1.1.1 Radiology reports

Radiology reports convey information about the clinical findings and the radiologist interpretation: accurate reporting of imaging is critical to patient care, as its content can influence both diagnosis and treatment decisions. Although SIRM has provided a definition [7], there is no universal consensus on what constitutes a clear and comprehensive radiology report: in general, this should include an exhaustive description of techniques, the key findings, the answer to the clinical query, and the radiologist's conclusions.

The European Society of Radiology (ESR) and the Radiological Society of North America (RSNA) have recognized the increasing need for the adoption of structured reporting (SR) in clinical practice [3,4]. SR is defined by Nobel et al. [8] as "an IT-based method to import and arrange the medical content into the radiological report". SR registries are digital documents in modular format, i.e. they use standardized templates oriented to the content and to the diagnostic query and are divided into sections (or levels) consisting of an ordered set of items, which require to be filled in, preferably and where possible, using standard terminology [3,9,10]. They have a significant role in advancing comprehensibility, standardization, completeness of information content, communication, data retrieval and large-scale data mining (e.g. diagnostic surveillance, disease classification, quality compliance, eligibility for clinical trials, epidemiological analysis) [3,9–12]; focusing on the quality, quantification and accessibility of information [13,14]. SR may improve workflow in the radiological and clinical routine (through its integration with existing tools and protocols, e.g. RIS/PACS infrastructures, free-text dictation etc.) by reducing inter-reader variability and reporting times compared to conventional reporting [3,9].

Despite general awareness, currently the diffusion and implementation of SR in clinical routine is limited, and most of the information is conveyed through conventional narrative reporting, hindering the rapid extraction, communication and use of clinical information by clinicians and healthcare systems. Indeed, a survey [9] conducted among Italian radiologists registered with SIRM showed that while SR is considered a desideratum, 56% of the interviewees never used it in their daily clinical practice. The main reasons are the data entry (which is time-consuming and labor-intensive when using mouse and keyboard), the lack of available commercial products for efficient and simple SR, the rigidity of the templates (e.g. complex clinical cases are sometimes difficult to categorize), and the low willingness to radically change personal work habits [9,15–17]. Improved standardization and automated AI-driven solutions may be helpful to promote SR adoption; it is thus evident that NLP techniques are key to transforming existing reports into a structured form, thus reducing human workload [2,12] while bringing all the advantages of SR.

### 1.1.2 Structured CT-based reports for Lymphoma patient staging

Lymphomas are blood tumors that result from genetic mutations in lymphocytes, which accumulate uncontrollably in the lymph nodes and circulate through the lymphatic and cardiovascular systems. This disease may affect not only lymphatic sites (the organs of the lymphatic system, i.e., the thymus and spleen, or a group of lymph nodes) but also extranodal sites (i.e., the lungs, liver, bone and spinal cord, kidneys, brain). A specialist can make the diagnosis and categorize lymphoma by reviewing the morphology, immunohistochemistry, flow cytometry and (if appropriate) molecular studies. A very important step is the staging of the disease, which suggests prognostic information and to keep disease progression monitored. Computed Tomography (CT) should be performed to obtain high quality images, accurate measurements and to discriminate between single large nodal masses and aggregations; it can allow the specialist to assess the extent, number and location of the lymph node stations affected by the tumor and, therefore, the stage [14].

A panel of experts of SIRM designed structured CT-based reports for the staging of patients with lymphoma. The final version of this SR registry (Granata et al. [14]) collects the clinical and radiological data of patients with lymphoma. It has four sections: *Patient Clinical Data*, *Clinical Evaluation, Imaging Protocol* and *Report*. As suggested by the radiologists, the entire analysis in this article focuses on the latter section, which includes the items related to the diagnosis, grouped into four macro groups defined by the "Site" of the lesion of the primary tumor *(Lymph node disease*, *Bulky disease*, *Spleen* and *Extranodal disease)*. To properly perform the staging phase, radiologists can take advantage of SR guiding them in fully describing the findings in the exam, thereby reducing reporting and data entry errors [14,16].

## 1.2   Research goals

In this article, we investigate the potential of NLP to transform previously written free-text radiology reports into a structured format, i.e., to extract from the reports the information needed to automatically fill in the items of the SR registry. Our pipeline relies upon generative, Transformer-based language models [18]: the current SoA in most of NLP tasks.

To achieve our goal, the following research questions are helpful to select and develop the most appropriate computational pipeline:

- RQ1: Can generative, Transformer-based language models help in filling radiology registries?
- RQ2: Radiology notes can be very long. What's the best strategy to address inputs exceeding language models context-size capabilities?

Finally, given the increasing popularity of large language models (LLMs), characterized by extremely high computational and economic costs:

- RQ3: Can smaller fine-tuned models achieve competitive performance while providing a more affordable solution for radiology SR registry filling?

## 1.3   Related works

AI has the potential to improve several aspects of the work performed by radiologists: planning, triage, interpretation of outcomes and their post-processing, imaging, reporting, clinical decision-making, etc. [2] NLP techniques are largely employed in radiology [6], e.g. to identify and extract concepts mentioned in reports, to classify reports or to generate structured outputs. The main areas of research concern diagnostic surveillance, cohort building for epidemiological studies, query-based case retrieval, quality assessment of radiology practice, clinical support services, etc. The literature offers many examples of biomedical text mining applications, which depend on the specific purpose and performance to be achieved, although sometimes using the same methodologies [19].

Previous analyses involving medical registry filling have focused on rule-based systems, such as those of Jorg et al. [15], Odisho et al. [20] and Sagheb et al. [21], based on regular expression and sometimes enhanced by ontologies and dictionaries. Esuli et al. [22] have relied on ML systems, while Viani et al. [23] on recurrent neural networks. Tavabi et al. [24] have proposed an interpretable classification approach by extracting sentences from clinical notes using Term Frequency–Inverse Document Frequency (TF-IDF). The transformer architecture, well known for the attention mechanism of Vaswani et al. [18], has been successfully applied to radiological reports, e.g., by Nowak et al. [25], Yan et al. [26] and Putelli et al. [27]; in another similar context, Buonocore et al. [28] explored a new rule-free extractive Question Answering (QA) approach based on BERT for cardiology registry filling in Italian.

Similarly, in this article, we want to solve the SR registry filling task without using rules and/or regular expressions, or any formal knowledge representation method, but rather by using a generative Transformer-based model: a scenario that has not yet been applied specifically to Italian radiology reports, and constitutes an original contribution of the present article.

## 2 Materials and Methods

The following analysis consists in examining the specific use case and the available dataset (Section **Error! Reference source not found.**), then finding the most appropriate model (Section **Error! Reference source not found.**) (RQ1) and designing a suitable pipeline (Section **Error! Reference source not found.**). To better analyze the problem at hand, two different strategies are compared (RQ2), and the best one is further compared to the most popular and cost-effective LLM at the time of writing, i.e., GPT-3.5 (RQ3).

### 2.1 Dataset

The dataset consists of 174 free-text radiology reports (in Italian language) on CT examinations investigating lymph-node lesions, supplied by the Department of Radiology, I.R.C.C.S. Policlinico San Matteo Foundation of Pavia, Italy. The reports were provided, subject to the approval of the ethics committee, completely anonymized and limited to the portion of text related to the techniques used during the CT examination, the anatomical areas involved, the radiological findings and the relevant observations and conclusions. Their length varies from a minimum of 152 words (408 tokens, in terms of minimum reading and writing units of text for language models) to a maximum of 1150 words (2729 tokens). Two radiology physicians from the same department (annotators) and with the same experience were asked to fill in a spreadsheet, shown in Figure 1, for each report (analogous to the SR registry), independently.

## Lymph Node Disease

| Number | Site | | | | Size | | Number of reference image | CT Appearance | | | | Relationship with neighboring structures | Complications | Notes |
|---|---|---|---|---|---|---|---|---|---|---|---|---|---|---|
| | Stage | Supradiaphragmatic sites | Subdiaphragmatic sites | Site | Largest dimension on axial plane (mm) | Dimension of the axis perpendicular to the largest diameter (mm) | | Structure | | | Other | | | |
| | | | | | | | | Areas of contrast enhancement | Areas of necrosis/ colliquation | | | | | |

## Bulky Disease

| Site | Largest Dimension Diameters (mm) | | | Reference Image | | | | CT Appearance | | Relationship with neighboring structures | | | Complications | Notes |
|---|---|---|---|---|---|---|---|---|---|---|---|---|---|---|
| | AP | LL | CC | Number of reference image | Diameters on reference image (mm) | | | Structure | | | | | | |
| | | | | | AP | LL | CC | Areas of contrast enhancement | Areas of necrosis/ colliquation | Airways | Vessels | Other | | |

## General Information

| Patient ID | Report ID | Site |
|---|---|---|
| | | |

## Spleen

| Largest dimension (measured on longitudinal MPR) | Structure | Focal Lesions | | | |
|---|---|---|---|---|---|
| | Structure | Number of target lesions | Largest dimension on axial plane (mm) | Dimension of the axis perpendicular to the largest diameter (mm) | Number of reference image |
| | | | | | |

*Figure 1 Spreadsheet provided to radiologists to collect annotations. Each entry (spreadsheet row) corresponds to a single report, identified by the General Information (gray fields in lower left corner).*

This represents the annotated dataset used to train the AI model later described. In this scenario, the verb "to annotate" means to complete the items of the spreadsheet without necessarily transcribing faithfully the words found in the text, as a radiologist would do in clinical practice, allowing permutation of terms and paraphrases. In order to assess the reliability and reproducibility of the work and the Inter Annotator Agreement (IAA), we deliberately allowed a fraction of overlapping reports (15% of the total amount of reports) among annotators. The IAA is on average about 0.7, falling within the range of "Substantial Agreement" (0.61-0.80) [29].

Missing values have been processed for each instance of the source dataset, distinguishing into two categories depending on the information they conveyed: (a) missing at random or undetected data, appearing in the annotations as "-" or "N.A." have been mapped into "NaN"; (b) fields that are intended to be empty (e.g., specifiers conditioned on the occurrence of given event, reported in previous items) have been mapped into "not applicable".

The items in the SR registry (36) cover a wide range of information, but only part of such items is sufficiently represented in the reports to be included in the training set. In particular, we identified a subset of features using the best trade-off between IAA and data completeness (i.e., smallest percentage of missing values) as a selection criterion. We hence focused on the subset "*Lymph node disease*" (14 items) resulting in six features, described in Table 1: three categorical features, one free-text and two continuous numerical features. Instances with "Site" equal to "Bulky" or "Spleen" were excluded.

The final dataset contains 1020 instances (340 for factual, 170 for free-text, 510 for multichoice).

## 2.2 AI model and NLP task selection

Our work leverages the Transformer architecture proposed by Vaswani et al. [18], an encoder-decoder structure relying entirely on the attention mechanism to represent input-output global dependencies. Transformer-based models are trained through a particular type of semi-supervised learning consisting of two phases: (1) pre-training, during which the model receives a large amount of unlabeled textual data as input and computes a general representation of it in an unsupervised manner; and (2) fine-tuning: supervised training using a limited set of domain-specific and labeled examples.

Different types of Transformer-based models have been proposed over the years, differing in structure and purpose: encoder-only models (e.g., BERT) used only the encoder block of the Transformer and are useful for input comprehension tasks (e.g., text classification, named entity recognition); decoder-only models (e.g., GPT) or encoder-decoder models (e.g., T5) leverage the decoder stack to solve text generation tasks such as machine translation, text summarization and question answering. The presence of free-text items, with rephrasing and reworkings, combined with the complexity and information density of input clinical notes, makes the encoder-decoder paradigm the best candidate for our SR registry filling task.

A popular encoder-decoder model is T5 (Text-to-Text Transfer Transformer), proposed by Raffel et al. [30], which has helped to advance the state of the art for many NLP tasks. Its architecture is particularly suitable for sequential tasks that cannot be performed by encoder-only models and can be laborious for decoder-only models due to the lack of explicit conditioning on the source context [31]. Among the key features of T5, there is the way it addresses NLP tasks using a text-to-text format, i.e. processing text as input and producing text as output; this enables the use of the same model, goal, training and decoding procedure for each task [32]. Although multilingual variants of the model have been introduced, they provide suboptimal performances for languages other than English compared to monolingual variants, like IT5, a monolingual version of T5 introduced by Sarti & Nissim [31] for the Italian language, released with different parameter sizes. For our study, we chose the computationally lightweight IT5 Base version (220M of parameters) and selected a generative Question Answering (QA) approach. The IT5 model for QA[1] has been trained on the SQuAD-IT dataset [33], which consists of a set of about 50000 (non-biomedical) Italian-specific paragraph-question-answers triplets.

---
[1] HuggingFace repository: it5/it5-base-question-answering

## 2.3 Experimental design

To test the model on our dataset, we draft a question for each feature selected from the SR registry, grouping them according to the type of answer required (as shown in Table 1): multichoice, free-text and factual.

*Table 1 Subdivision of features and related questions in three groups according to the type of answer they require: multichoice, free-text and factual. English translation for the Question fields, from top to bottom: Is only lymph node disease diagnosed? How many lymph node stations does the tumor affect? What stage is lymphoma? Where are the lymph node stations affected by the tumor? What is the largest axial dimension of the primary tumor? What is the dimension of the axis perpendicular to the maximum diameter of the primary tumor?*

| Group and characteristics | Feature | Question | Answer |
|---|---|---|---|
| **Multichoice:** categorical response bound to a list of options | Lymph only (categorical) | *La diagnosi è solo di malattia dei linfonodi?* | True, False |
| | Number of stations (categorical) | *Quante sono le stazioni linfonodali interessate dal tumore?* | 1, 2+ |
| | Stage (categorical) | *Qual è lo stadio del linfoma?* | Limited disease (stage I-II), Advanced disease (stage III-IV) |
| **Free-text:** unbound response (e.g., a set of terms or sentences) | Site (free-text) | *Dove si trovano le stazioni linfonodali interessate dal tumore?* | Free-text |
| **Factual:** semi-bound numerical response | Axial plane size (continuous) | *Quanto è la dimensione massima sul piano assiale del tumore primario?* | *Number* mm |
| | Perpendicular axis size (continuous) | *Quanto è la dimensione dell'asse perpendicolare al diametro massimo del tumore primario?* | *Number* mm |

We concatenate information from each radiology report with the question and allowed options where possible, labeling it as *input-text*, while labeling the correct answer as *output-text* (as reported in Figure 2).

Since the average length of radiology reports (1220 tokens) exceeds the maximum input text length allowed by T5 (512 tokens), two different processing strategies are employed and compared:

(1) ***Batch-truncation***: it consists of considering only the first batch (i.e., first 512 tokens) of each report;
(2) ***Ex-post combination***: it consists of using the texts integrally by splitting them into 512-token-long batches and then combining the best outputs per report downstream of obtaining the results (after two sequential validations).

For both the strategies, we implement stratified 5-fold cross-validation (CV) to fine tune the model and evaluate its performance. Preliminary analyses have shown biased results for the multichoice type due to the presence of a significantly unbalanced feature (i.e., Number of stations (1/2+)), which we fixed by under-sampling the majority class. For *ex-post combination* strategy, we have tuned some text-generation hyperparameters on each validation set to improve the results of free-text answers (see Supplementary Materials Section 1).

When using generative language models for structured reporting, it is important to control text generation to ensure consistency with the standardized expected output. For this reason, two generation techniques are implemented: the first, called *unconstrained generation*, lets the model generate freely; the second, called *constrained generation*, uses constrained beam search [34] to reduce the combinations of tokens that can be explored during generation to a finite set of allowed sequences (i.e., a whitelist of possible answers), preventing forbidden answers from being generated. For multichoice items, the whitelist is populated with the possible values each item can assume; for factual items, like measurements, the whitelist includes only the tokens included in the original text; for free-text items, *constrained generation* is not used since it's not possible to define any whitelist a priori.

To confirm the capabilities of the optimized IT5 model in extracting information for SR registry filling task, we compare the best IT5 configuration with the most capable and cost-effective large-scale model available at the time of designing this work, i.e., GPT 3.5 Turbo, which is several orders of magnitude bigger and supports much longer input texts. Since *constrained generation* requires direct manipulation of the logits, which are not accessible in closed-source environments like OpenAI's GPT, we leverage prompt engineering techniques to design ad-hoc prompts to better guide the model in performing the SR registry filling task.

Quantitative evaluation metrics are coupled with human-expert feedback by providing radiologists a questionnaire [35–37] about the quality of the free-text generated answers to assess which model reflects better the way radiologists fill structured reports. Specifically, the degree of completeness and correctness are evaluated using two independent 5-point ordinal Likert scales (more details are reported in Supplementary materials Section 2). Finally, statistical analyses are performed to compare the scores.

An overview of the overall analysis is shown in Figure 2. The source code for the experimental setup is available on github[2].

---

[2] https://github.com/bmi-labmedinfo/nlp-radiology-paper.git

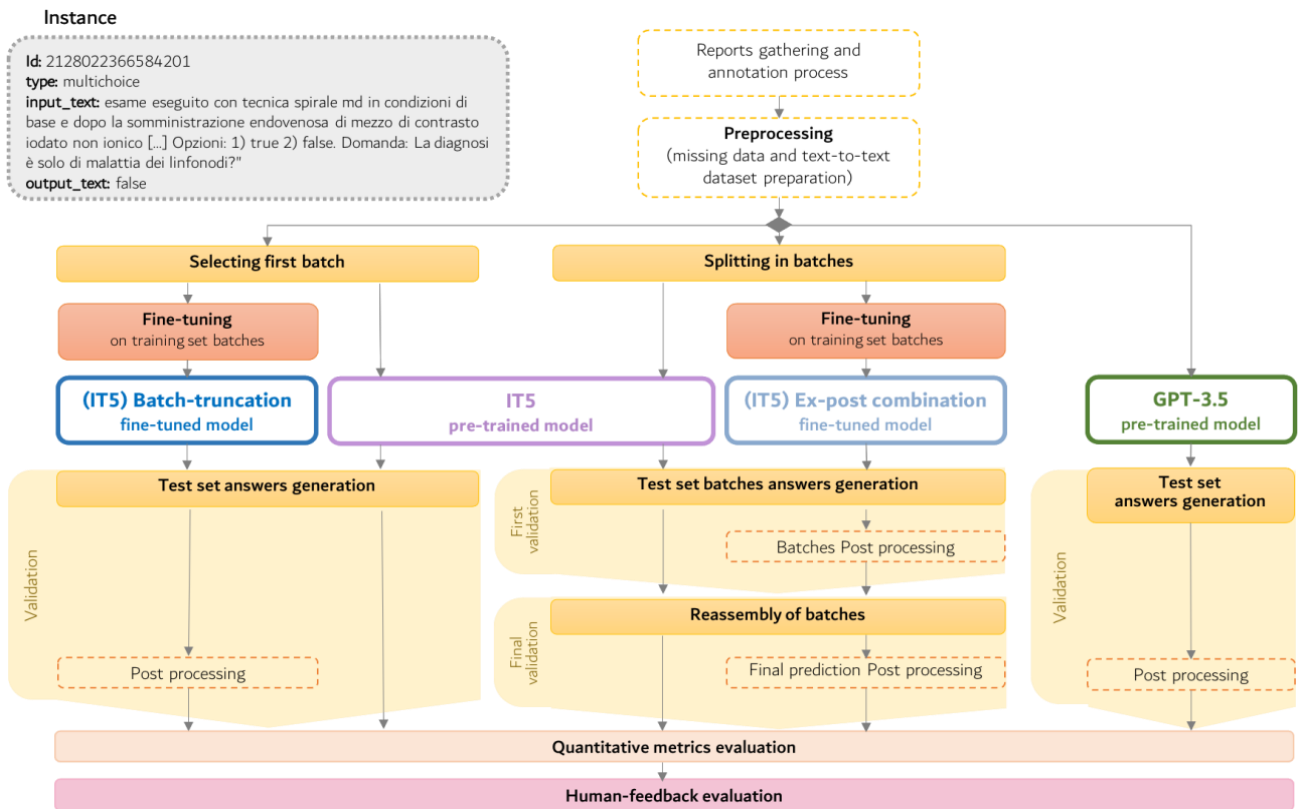

*Figure 2 SR registry filling pipeline overview. The data gathering is designed to collect CT reports with radiological findings that can be classified as Lymph node disease, Bulky disease, Spleen or combinations thereof, i.e., falling within the macro "Site" of the SR registry. The annotations of the reports are preprocessed: first, missing data are replaced by "Nan" or "not applicable"; second, annotations not belonging to the selected feature subset are dropped. Each instance is transformed as input_text and output_text, as shown in the gray box (English translation for input_text: examination performed with md spiral technique under baseline conditions and after intravenous administration of non-ionic iodinated contrast [...] Options: 1) true 2) false. Question: Is only lymph node disease diagnosed?)*

*The following steps are repeated for each fold of the 5-fold cross-validation. A model is fine-tuned on the training set for IT5 strategies; pre-trained and fine-tuned models are tested on the test set and their results go through a validation phase where quantitative metrics results are collected and compared to human-expert feedback ratings. Post-processing steps refer free-text answers and involve the removing of truncated words, repeated sentences, "not applicable", and final punctuation.*

*The validation phase for ex-post combination is divided in two steps: first, all the batches are validated with respect to their output_text, after which batches are grouped by id and reassembled differently according to the type of the answer (batch with higher confidence for multichoice and factual, concatenation of batches with confidence higher than the empirical threshold of 0.2 for free-text, or just a single batch when confidence is 1). Then, the resulting reassembled answers go through the final validation process, in the same way as batch-truncation and GPT-3.5.*

## 2.4 Evaluation metrics

To quantify and track the performance during fine-tuning, we compute the ROUGE metric [38] (in its Longest Common Sequence variant), BERTScore [31,39], F1-score and loss on the dev set, using the latter as an early stopping criterion, saving the best checkpoint for each fold. The performance over final answers obtained by each strategy and model is reported comparing generated answers and annotators' answers in terms of strict accuracy (SA), F1 and format accuracy (FA). In this context, F1 computes the tokens in common (considering Precision as the ratio of the number of common tokens over the total number of predicted tokens, while Recall as the ratio of common tokens over the total number of true answer's tokens), SA refers to the percentage of correct answers (i.e., exact match), and FA assesses whether the format of the predicted answer coincides with the expected one (e.g., size of the tumor reported as an integer number followed by the "mm" unit of measurement).

# 3 Results

Table 2, Table 3 and Table 4 report the overall (as micro-average over all types of answer in CV folds) and type-wise average results of the validation of the models performed on test sets. For each strategy, the best results (by type and overall) are highlighted in bold.

*Table 2 Overall and type-wise average results (with standard deviation) of IT5 batch-truncation.*

| type | IT5: Batch-truncation | | | | | | | | | | | |
|---|---|---|---|---|---|---|---|---|---|---|---|---|
| | Pre-trained model | | | | | | Fine-tuned model | | | | | |
| | unconstrained | | | constrained | | | unconstrained | | | constrained | | |
| | SA | F1 | FA | SA | F1 | FA | SA | F1 | FA | SA | F1 | FA |
| *Factual* | 0.6 ±0.8 | 51.5 ±2.3 | 19.7 ±4.5 | 0.6 ±0.8 | 51.4 ±2.3 | 18.8 ±4.5 | **17.1** ±3.5 | **68.8** ±2.8 | **85.6** ±2.8 | 2.4 ±2.9 | 52.9 ±3 | 30 ±5.8 |
| *Multichoice* | 0.3 ±0.6 | 26.4 ±2.4 | 1.1 ±1.1 | 46.2 ±6.3 | 74.5 ±3.8 | **100** ±0 | 57.1 ±6.7 | 75.1 ±7 | 100 ±0 | 57.1 ±6.7 | 75.1 ±7 | 100 ±0 |
| *Free-text* | 0 ±0 | 16.3 ±2.1 | **100** ±0 | - | - | - | **27.8** ±9.8 | **39** ±10 | 100 ±0 | - | - | - |
| *Overall* | 0.3 ±0.3 | 34.2 ±1.9 | 27.4 ±1.8 | 19.5 ±2.7 | 54.3 ±2.5 | 68.5 ±1.7 | **35.9** ±2.5 | **65.7** ±2.7 | **94.4** ±1.1 | 30.2 ±2.3 | 59.5 ±2.5 | 72.8 ±2.2 |

*Table 3 Overall and type-wise average results (with standard deviation) of IT5 ex-post combination. \* indicates the very best results among all models.*

| type | IT5: Ex-post combination | | | | | | | | | | | |
|---|---|---|---|---|---|---|---|---|---|---|---|---|
| | Pre-trained model | | | | | | Fine-tuned model | | | | | |
| | unconstrained | | | constrained | | | unconstrained | | | constrained | | |
| | SA | F1 | FA | SA | F1 | FA | SA | F1 | FA | SA | F1 | FA |
| *Factual* | 1.2 ±1.2 | 63.8 ±2.2 | 47.9 ±4.2 | 1.5 ±1.8 | 63.3 ±2.1 | 45.3 ±3.9 | **46.5\*** ±10.4 | **87.1\*** ±2.9 | **88.2\*** ±5.2 | 13.2 ±4.3 | 70.3 ±2.6 | 50.3 ±8.9 |
| *Multichoice* | 1.6 ±1.8 | 34.3 ±3 | 2.2 ±2.1 | 51.3 ±5.7 | 75.6 ±3.5 | **100** ±0 | **64.7\*** ±9.2 | **78.1\*** ±10.1 | **100\*** ±0 | 64.7 ±9.2 | 78.1 ±10.1 | 100 ±0 |
| *Free-text* | 0 ±0 | 26.7 ±2.8 | **100** ±0 | - | - | - | **33.7\*** ±12.6 | **56.3\*** ±12.5 | **100\*** ±0 | - | - | - |
| *Overall* | 1.1 ±0.7 | 44.3 ±2.5 | 38.9 ±2.1 | 22.1 ±2.2 | 61.4 ±2.6 | 78.7 ±1.5 | **51.7\*** ±6.7 | **77.4\*** ±6.2 | **95.4\*** ±2 | 38.7 ±5.7 | 70.9 ±5.4 | 80.7 ±3.5 |

Table 4 Overall and type-wise average results (with standard deviation) of GPT-3.5.

| type | GPT-3.5 | | | | | | | | | | | |
|---|---|---|---|---|---|---|---|---|---|---|---|---|
| | Pre-trained model | | | | | | Fine-tuned model | | | | | |
| | unconstrained | | | constrained | | | unconstrained | | | constrained | | |
| | SA | F1 | FA | SA | F1 | FA | SA | F1 | FA | SA | F1 | FA |
| *Factual* | **5.9** ±2.3 | **17.5** ±3.2 | **29.4** ±5.6 | N.A. | N.A. | N.A. | N.A. | N.A. | N.A. | N.A. | N.A. | N.A. |
| *Multichoice* | **54.4** ±2.9 | **64.4** ±2.7 | **6.3** ±2.3 | | | | | | | | | |
| *Free-text* | **2.4** ±2.5 | **33.8** ±9.6 | **100** ±0 | | | | | | | | | |
| *Overall* | **25.5** ±1 | **40.3** ±2.2 | **33.4** ±2.4 | | | | | | | | | |

For IT5 models (Table 2 and Table 3), the *constrained generation* (*constrained* columns) improves only multichoice and overall pre-trained models results (at the expense of an increase in the dispersion of results for SA and F1); the best results (in bold) gather in the fine-tuned *unconstrained generation* condition.

For this reason, from now on the terms *batch-truncation* and *ex-post combination* refers exclusively to the fine-tuned and *unconstrained generation* conditions. Of these two, *ex-post combination* turns out to be the best strategy, even outperforming GPT-3.5 (as shown in Figure 3).

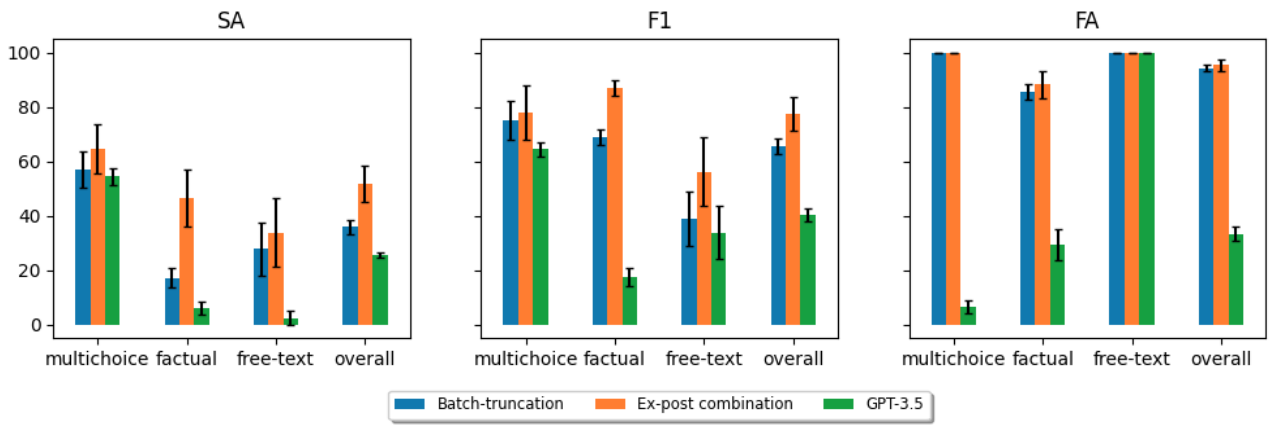

Figure 3 Comparison between the procedures (with standard deviation) divided by metric.

## 3.1 Comparative analysis of model performances

Paired t-test and Wilcoxon signed-rank test have been used on 5-fold CV metrics results to check for statistically significant differences between the two IT5 strategies, and to compare *ex-post combination* (best candidate) with GPT-3.5. The resulting p-values are shown in Figure 4.

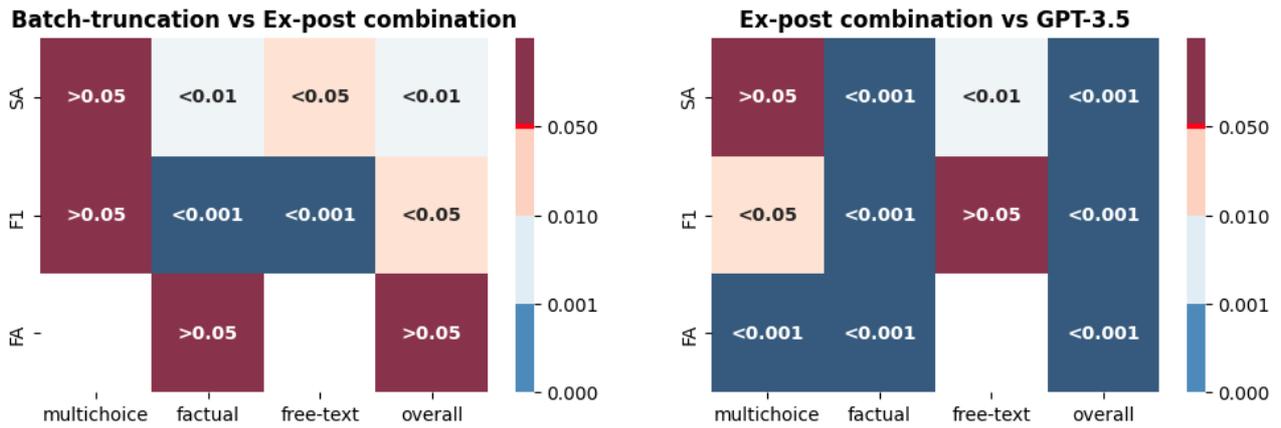

*Figure 4 Heat Maps of the p-values obtained by the comparison between batch-truncation and ex-post combination (on the left) and between ex-post combination and GPT-3.5 (on the right). The red mark on the colorbar indicates the significance threshold (0.05) for p-values: below this threshold the differences are statistically significant, while above (burgundy cells) the differences are non-significant; white (with no value) indicates that the p-value could not be computed because the differences between the metrics were all equal to zero.*

The comparison between *batch-truncation* and *ex-post combination* shows a not statistically significant difference between the two strategies for $SA_{multichoice}$ (p-value=0.27) and $F1_{multichoice}$ (p-value=0.61), while for *ex-post combination* and GPT-3.5 differences are not statistically significant for $SA_{multichoice}$ (p-value=0.081) and $F1_{free\text{-}text}$ (p-value=0.062).

Regarding the multichoice type, in both comparisons the similarities are in the generation of responses for the "Stage" (limited/advanced) and "Number of stations" (1/2+) features (as shown in Figure 5).

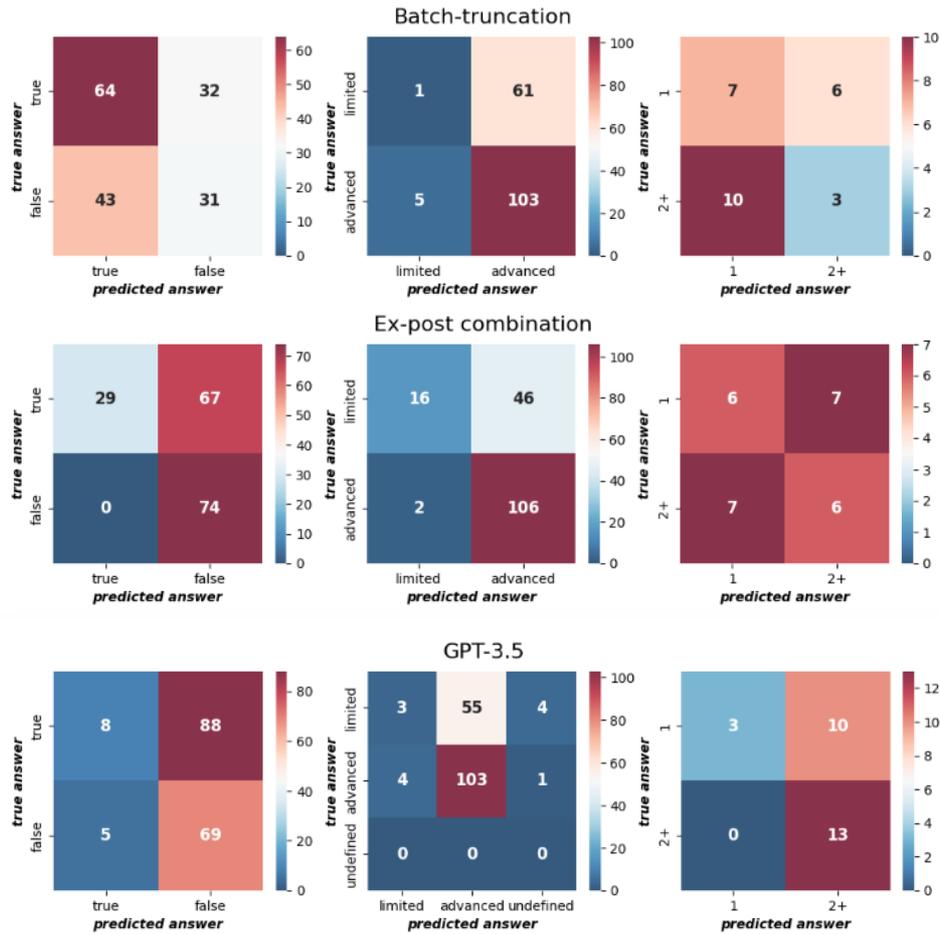

*Figure 5 Confusion matrices showing the number of correct and incorrect matches for the multichoice type. For each model, matrices refer to "Lymph only" (true/false), "Stage" (limited/advanced) and "Number of stations" (1/2+) features, respectively. For the "Stage" feature, GPT-3.5 sometimes is unable to give an answer, so the label "undefined" has been added to group all predicted answers other than limited/advanced.*

Regarding the free-text type, the formal way in which the F1 metric is defined does not consider whether the model is able to properly answer with "not applicable" when actually appropriate (57 out of 62 for *ex-post combination*, while 4 out of 62 for GPT-3.5), as shown in Figure 6; instead, it evaluates the similarity between the predicted and true answer and is thus driven by "narrative" answers (which all differ from "not applicable"). Further in-depth considerations are discussed in the Section 4.

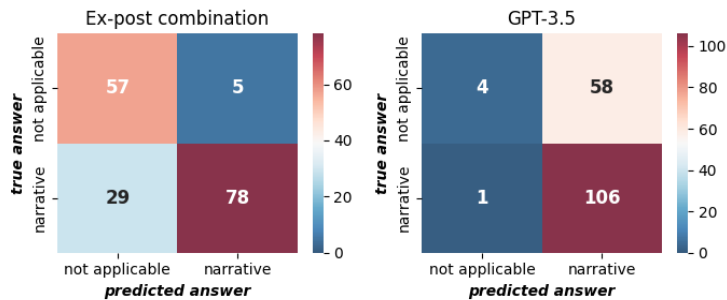

*Figure 6 Confusion matrices showing the number of correct and incorrect matches for ex-post combination and GPT-3.5; the label "narrative" groups all the free-text answers other than "not applicable". It is important to note that "narrative" matches do not suggest that the answers coincide; rather, they indicate that both the predicted and true answer differ from "not applicable".*

## 3.2 Human-expert based evaluation

The evaluation of models' answers by the medical experts, comparing for IT5 *ex-post combination* and GPT-3.5, achieves Cohen's kappa values <0.3 (considering all scores together and split per model) for IAA, which falls within the range of "Fair Agreement" (0.21-0.40) [29]. Distributions of Likert-scale answers are shown in Figure 7.

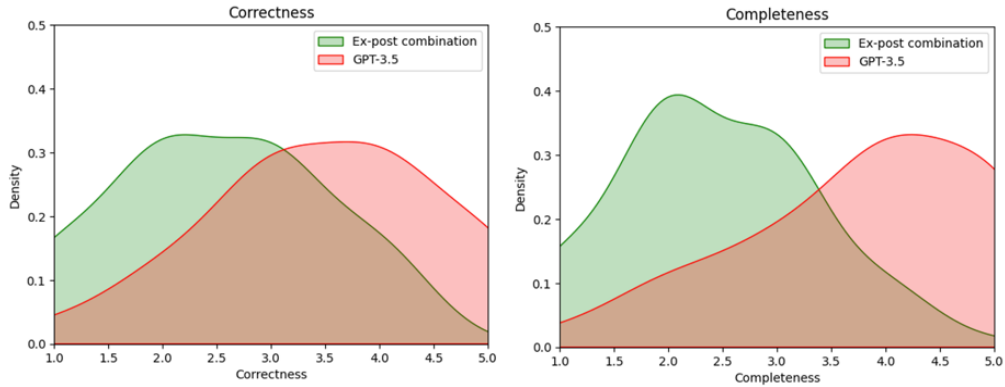

*Figure 7 Distribution of Likert-scale answers (from 1 to 5) grouped by evaluation criterion (correctness on the left and completeness on the right) and model (colors). For both criteria, the distributions related to GPT-3.5 (correctness: 3.50±1.6 and completeness: 3.81±1.10) have a higher central tendency than IT5 ex-post combination (correctness: 2.51±0.97 and completeness: 2.43±0.90).*

The differences between the common answers are tested for normality by the Shapiro-Wilk test, reporting correctness and completeness p-values<0.001 (i.e., they do not follow a normal distribution). Therefore, we use the non-parametric Wilcoxon signed-rank test that shows a statistically significant difference (correctness and completeness p-values<0.001). Central tendencies suggest that GPT-3.5 answers have higher scores than *ex-post combination*.

We check the correlation between the two independent criteria and between them and F1 by computing Spearman's correlation coefficient with the corresponding significance test. A significant correlation exists between the completeness and correctness criteria (p-values<0.001), and also with F1 (p-values<0.001) for both models.

## 4 Discussion

The results we obtained for IT5 strategies, especially for *ex-post combination*, suggest that batch partitioning has good potential (even without fine-tuning the model using domain-specific training examples, i.e., utilizing pre-trained only models): inevitably, *batch truncation* allows the model to have awareness of only a severely limited portion of the text, where the desired response is likely to be contained only partially or not at all (e.g. the free-text item requires listing the location of the lymph node lesions, which are scattered throughout the text of the report), resulting in sub-optimal performance.

The comparison of these strategies shows very similar behavior in generating multichoice answers; however, the two models differ in approaching responses related to the "Lymph only" feature (as shown in Figure 5). When the disease affects only lymph nodes (correct answer=*true*), the information often stands in the first batch and *batch-truncation* answers correctly (64 out of 96); while the other batches carry a lot of noisy and unnecessary information that mislead ex-post combination (29 correct out of 96). On the other hand, when the correct answer=*false*, *ex-post combination* always answers correctly (74 out of 74), while *batch-truncation* does not have enough information to answer *false* (31 correct out of 74).

The work aims to create a pipeline that can extract data from reports with different writing styles and document structures; for this reason, the model should have a view of the entire report, ignoring the fact that the information may be summarized in the first or last sentences of the text, confirming that the *ex-post combination* is the better, and more generalizable, of the two strategies.

Regarding GPT-3.5, suboptimal values can be observed for the factual type mainly due to the inability of the model to find the answer in the text, as well as unit-of-measurement or more-than-one-size errors (e.g., the model predicted "16x27 mm", while the correct answer was "16 mm"). The comparison with *ex-post combination* shows a similar behavior in generating free-text answers; we can observe that *ex-post combination* has a clear advantage over GPT-3.5 with regards to the ability to appropriately output "not applicable" as an answer (57 out of 62 correctly answered as "not applicable by IT5 *ex-post combination*, vs. 4 out of 62 by GPT-3.5), as shown in Figure 6. On the other hand, GPT-3.5 gives more plausible answers (as highlighted by the human-expert-based evaluation, see Section 3.2) that result in a higher F1 score, but always attempt to provide an answer (i.e., different from "not applicable") in any case, whether appropriate or not given the information included in the original free-text report.

The human-expert based evaluation highlights the difficulty in objectively, and uniformly, judging the correctness and completeness of a sentence; this is demonstrated by low IAA values. Table 5 shows some examples of ratings given by the annotators: they show how (in those cases where *reference* and *model answer* are very similar) *annotator 1* tends to give lower scores while *annotator 2* gives the maximum to both criteria.

*Table 5 Examples of human-expert-based evaluation.*

| Reference | Model answer | Annotator 1 | Annotator 2 |
|---|---|---|---|
| Latero-cervical, axillary, retro-pectoral, mediastinal, mammary, pulmonary hilum, paratracheal, epiphrenic, intraperitoneal, retroperitoneal | Latero-cervical, axillary/retro-pectoral, mediastinal (mammary, paratracheal, of the lung hilum, posterior mediastinum, and epiphrenic), subcarinal, intra- and retroperitoneal stations | Completeness: 3 Correctness: 3 | Completeness: 5 Correctness: 5 |
| Later-cervically on the left side, at the upper thoracic inlet, paratracheal and periesophageal bilaterally, mesenteric | Latero-cervical regions, upper thoracic inlet, paratracheal, periesophageal, mesenteric (left quadrants) | Completeness: 4 Correctness: 2 | Completeness: 5 Correctness: 5 |

In general, GPT-3.5 shows a remarkable ability to generate plausible statements, resulting in satisfactorily evaluated answers by physicians, most likely thanks to its 175B of parameters allowing it to output very realistic, human-like, text.

The positive correlation between the criteria and F1 suggests that the latter is a good computational approximation of what a clinician would rate.

The original contribution of this paper concerns the use of generative Transformer-based models to deal with the extraction of information of different types. Indeed, we aim to use the QA approach to generate categorical (multichoice), numerical, and free-text answers: related works (see Section 1.3) focus on a single type of data (i.e., our multichoice type).

Our study also shows several limitations. Lower performance is observed in generating responses with type free-text; indeed, these items contain a large amount of information, scattered throughout the report and mentioned with varying degrees of significance. Another limitation is the collection of reports, both in terms of number and content. Although considerable time and effort have been devoted to the annotation of the corpus, the size of the dataset is small; by increasing the number and heterogeneity of examples, new challenges could question the capabilities of the system, deserving further follow-up investigation and evaluation of performances.

The richness of details and information content is not uniform in all the reports in our dataset: this contributes to significant discrepancies between notes, preventing all items of the SR registry from being successfully tackled by our proposed approach for SR registry filling. To train the models with high-quality examples, we identified the most frequently filled items in the annotations, i.e., reducing the number of features analyzed and, consequently, limiting the information to be extracted.

Further more comprehensive studies are essential to gain a deeper understanding of the potential impact (e.g., the usefulness, the reproducibility, and the scalability) of the proposed pipeline. Considering more structurally different reports and features, up to a full coverage of the SR registry sections and fields, is a challenging task that we aim to address in follow-up research. Another important aspect to consider is time savings: we will ask clinicians how long it takes them to fill out the entire SR registry, using slots with an increasing number of reports, and compare that to the time they employ when supported by our proposed pipeline. Solutions prioritizing speed and precision are likely to be optimal for time-critical tasks and positively impact time spent on clerical and documentation tasks, ultimately freeing healthcare professionals time to be dedicated to the actual delivery of care.

## 5 Conclusion

In the biomedical domain, the main challenges related to cNLP depend on the complexity of the task (e.g., text classification, entity mapping, summarization, QA, etc.) and the characteristics of the texts to be analyzed and queried (heterogeneity and corpus size, information content, event definition). This article proposes an SR registry filling pipeline for the application of Transformer-based models in healthcare to transform previously written free-text reports into a specialized (radiological) SR registry format.

Our experiment shows that IT5 QA is a viable option to positively answer our RQ1 (*Can generative, Transformer-based language models help in filling radiology registries?*), while RQ2 (*What's the best strategy to address inputs exceeding language models context-size capabilities?*) is addressed by implementing *batch-truncation* and *ex-post combination*, with the latter showing better results. Regarding RQ3 (*Can smaller fine-*

*tuned models achieve competitive performance while providing a more affordable solution for radiology RS registry filling?*), *ex-post combination* achieves notable results, especially learning when to give a non-N.A. answer, even if showing lower performances in plausibility, according to a human-expert-based evaluation, when compared to GPT-3.5.

For this experimental setting, the results suggest a greater impact of in-domain data fine-tuning; IT5 performs well in addressing the SR registry filling task, even with a relatively small fine-tuning dataset.

Despite the available data, we aimed to create a pipeline that can extract data from reports with different writing styles and document structures. For this reason, the *ex-post combination* strategy outperforms *batch-truncation*, especially thanks to a higher degree of generalization and the view of the entire input text.

Although the models have very different numbers of parameters, the comparison between *ex-post combination* (220M of parameters) and GPT-3.5 (175B of parameters) does not show statistically different performance for all the types of answers (especially for multichoice and free-text field types). However, we can observe that the smaller fine-tuned model (IT5) learns to understand when a question needs to be actually answered, a considerable limitation shown by the pre-trained GPT-3.5 model which, on the other hand, tends to provide answers that are more liked by human expert evaluators.

## CRediT authorship contribution statement

**Laura Bergomi:** Conceptualization, Methodology, Software, Validation, Investigation, Formal analysis, Data Curation, Writing – Original draft, Writing - Review & Editing, Visualization. **Tommaso M. Buonocore:** Conceptualization, Methodology, Software, Validation, Investigation, Formal analysis, Resources, Data Curation, Writing – Original draft, Writing - Review & Editing, Visualization, Supervision. **Paolo Antonazzo:** Validation, Investigation, Data Curation. **Lorenzo Alberghi:** Validation, Investigation, Data Curation. **Riccardo Bellazzi:** Conceptualization, Supervision, Project administration. **Lorenzo Preda:** Conceptualization, Supervision, Project administration. **Chandra Bortolotto:** Conceptualization, Investigation, Supervision, Project administration. **Enea Parimbelli:** Conceptualization, Methodology, Software, Validation, Investigation, Formal analysis, Resources, Data Curation, Writing – Original draft, Writing - Review & Editing, Visualization, Supervision, Project administration.

## Declaration of competing interest

The authors declare that they have no known competing financial interests or personal relationships that could have appeared to influence the work reported in this paper.

## Appendix A. Supplementary data

Source code available at https://github.com/bmi-labmedinfo/nlp-radiology-paper.git.

**Highlights**

- Leveraging NLP and generative Transformer-based models for SR registry filling
- Rule-free QA approach deals with different types of variables (including free-text)
- Greater impact of in-domain data fine-tuning combined with batch partitioning
- Quantitative and qualitative (human-expert-based) evaluation of model answers
- Small fine-tuned model ensures answer's appropriateness and fair plausibility

**Supplementary materials**

## 1  Hyperparameters tuning

During *ex-post combination* fine-tuning, we tuned the text-generation hyperparameters shown in S. Table 1 on dev sets. The hyperparameters considered in the analysis (except *confidence threshold*) were chosen from those configurable in the *GenerationConfig* class of the Hugging Face *transformers* library.

*S. Table 1 Text-generation hyperparameters tuned during ex-post combination fine-tuning.*

| Grid search range | repetition_penalty | num_beams | length_penalty | confidence threshold |
|---|---|---|---|---|
| min value | 1 | 1 | 1 | 0.09 |
| max value | 1.6 | 4 | 3 | 0.4 |

We applied grid search to test each combination of hyperparameters (the ranges are shown in S. Table 1 as *min value* and *max value*). For each CV fold, the combination of hyperparameters with the maximum f1 and strict-accuracy values was used to calculate the mean values of the hyperparameters.

Average hyperparameters (*repetition_penalty=1, num_beams=4, length_penalty=2, confidence threshold=0.2*) were used in the subsequent generation phase on the test sets.

## 2  Human-expert based evaluation

We designed an instrument to collect human-expert feedback about the quality of the free-text answers generated by the models: specifically, on the degree of (i) completeness and (ii) correctness, using two independent 5-point ordinal Likert scales.

We created a spreadsheet (shown in S. Figure 1) contained, for each entry, the *id*, the *reference* annotation, the *model answer* to evaluate, two drop-down menus to be completed with a response bound from 1 to 5 (the description of the scores is given in S. Table 2) for the criteria (i) and (ii) and an extra field for any observation.

| ID | Reference | Model answer | Completeness | Correctness | Notes |
|---|---|---|---|---|---|
|  |  |  |  |  |  |

*S. Figure 1 Spreadsheet provided to radiologists to collect human-expert feedback.*

The evaluation process was as follows: two radiology physicians (annotators) with the same expertise were asked to complete the spreadsheet, without consulting, by comparing each *model answer* with the corresponding *reference* and linking a score (1-5) for each criterion (i, ii). The experts were blinded to which model produced each answer by randomly sorting them.

Only answers generated by fine-tuned unconstrained *ex-post combination* and GPT-3.5, whose reference and prediction were different from "not applicable", were included, for a total of 205 entries (98 and 107 respectively). Finally, statistical analyses were performed to compare the scores of the two models (specifically the individual ratings randomly selected from each pair of raters' evaluations) and evaluate their performance.

*S. Table 2 Definition of 5-points Likert scales used in human-expert based evaluation.*

| Score | Completeness: *How do you rate the completeness/exhaustiveness of the answer generated by the model according to the following criterion: have all the sites of the lymph node stations involved in the disease been reported?* | Correctness *How do you rate the correctness/accuracy of the answer generated by the model?* |
|---|---|---|
| 1- Unacceptable | The answer significantly lacks in the particular criterion; too few sites are mentioned. | The answer is fundamentally incorrect. |
| 2- Poor | The answer lacks in the criterion but not to a severe extent; few sites are mentioned. | The answer has evident inaccuracies. |
| 3- Fair | The answer adequately but not exceptionally meets the criterion; enough sites are mentioned. | The answer contains several inaccuracies that may require clarification or verification by a medical professional. |
| 4- Good | The answer aligns well with the criterion; most sites are mentioned. | The answer is mostly accurate, with only minor discrepancies that do not significantly impact its clinical reliability. |
| 5- Excellent | The answer excels in the criterion; all sited are mentioned. | The answer is thoroughly accurate, aligning perfectly with clinical knowledge |